# PVTAdpNet: Polyp Segmentation using Pyramid vision transformer with a novel Adapter block

Arshia Yousefi Nezhad, Helia Aghaei, Hedieh Sajedi *

Department of Mathematics, Statistics and Computer Science, College of Science,
University of Tehran, Tehran, Iran,
Corresponding author: Hedieh Sajedi, hhsajedi@ut.ac.ir

**Abstract**

Colorectal cancer ranks among the most common and deadly cancers, emphasizing the need for effective early detection and treatment. To address the limitations of traditional colonoscopy, including high miss rates due to polyp variability, we introduce the Pyramid Vision Transformer Adapter Residual Network (PVTAdpNet). This model integrates a U-Net-style encoder-decoder structure with a Pyramid Vision Transformer backbone, novel residual blocks, and adapter-based skip connections. The design enhances feature extraction, dense prediction, and gradient flow, supported by squeeze-and-excitation attention for improved channel-wise feature refinement. PVTAdpNet achieves real-time, accurate polyp segmentation, demonstrating superior performance on benchmark datasets with high mDice and mIoU scores, making it highly suitable for clinical applications. PVTAdpNet obtains a high Dice coefficient of 0.8851 and a mean Intersection over Union (mIoU) of 0.8167 on out-of-distribution polyp datasets. Evaluation of the PolypGen dataset demonstrates PVTAdpNet's capability for real-time, accurate performance within familiar distributions. The source code of our network is available at https://github.com/ayousefinejad/PVTAdpNet.git

1. **Introduction**

Colorectal cancer is a type of cancer that starts either in the colon or the rectum. It develops mainly due to the abnormal growth of healthy cells in the colon or rectum, resulting in polyps. With time, many of these polyps could become potential cancerous cells.
It is estimated that over 153 thousand new cases of colorectal cancer and 53 thousand deaths occurred in the United States, about 1 in 10 cases and deaths (Table 1) [1]. In aggregate, ranking third by incidence, and second by mortality, is colorectal cancer [2]. A marked heterogeneity in the incidence and mortality rates of colorectal cancer was evident across different geographic regions. Notably, Europe and Australasia exhibited the highest incidence rates, whereas Eastern Europe bore a disproportionate burden of mortality. Projections indicate a substantial escalation in the global colorectal cancer burden by 2040, with a projected 63% increase in new cases to reach 3.2 million annually and a 73% increase in mortality to 1.6 million deaths per year.

Colonoscopy is widely recognized as the most reliable method for diagnosing colorectal cancer. By enabling the direct visualization of the colon, colonoscopy facilitates the early detection and removal of polyps, which are precancerous growths that can develop into colon cancer if left untreated. Recent advances in deep learning technology are embracing promising solutions to improve diagnostic accuracy by highlighting the possible precancerous tissues and easing the workload of healthcare professionals [3, 4].

Deep learning has revolutionized computer vision, particularly in image segmentation. Architectures like U-Net and Fully Convolutional Networks (FCNs), employing encoder-decoder structures, have become foundational. These models progressively downsample images to extract meaningful features and subsequently upsample to refine localization. Skip connections enhance precision by preserving fine-grained details.

Attention mechanisms have emerged as a powerful tool in this domain. By selectively focusing on salient image regions, attention improves feature representation and obviates the need for redundant multi-scale features. This has led to significant advancements in tasks such as object detection, image classification, and, crucially, image segmentation.

This study presents PVTAdpNet, a transformer with an adapter layer that has been developed for precise polyp segmentation in medical imagery. PVTAdpNet incorporates advanced techniques to enhance polyp segmentation, resulting in improved accuracy and precision. It provides accurate results and efficient calculations, making it perfect for clinical applications. Here are the main highlights:

1. PVTAdpNet introduces an innovative residual decoder architecture to improve the gradient flow and facilitate multi-scale feature fusion. This enhances the model's capacity to locate polyps in complex medical images accurately.

2. PVTAdpNet achieves a higher benchmark in polyp segmentation, showcasing outstanding performance on widely recognized benchmark datasets while also pre-serving computational efficiency.
3. The adapter layer of PVTAdpNet, due to its projection, non-linearity, and up-projection components, enhances the capability for dense prediction and thus empowers the model to become stronger and more expressive of complex patterns.
4. PVTAdpNet shows a Mean Dice Coefficient (mDice) improvement of approximately 5.6% on the Kvasir-SEG dataset, 2.36% on the CVC-ClincDB dataset, and an average of 9.38% on the PolypGen dataset. Similarly, PVTAdpNet demonstrates a mIoU improvement of approximately 7.1% on the Kvasir-SEG dataset, 0.77% on the CVC-ClinicDB dataset, and an average of 9.28% on the PolypGen dataset.

The paper proceeds as follows. Section 2 establishes the groundwork by reviewing relevant prior research. Section 3 introduces the PVTAdpNet architecture in detail. Section 4 outlines the training methodology employed. Comprehensive experimental results are presented and analyzed in sections 5 and 6, respectively. The paper concludes with a summary of the findings in Section 7.

**TABLE 1** Estimated new cancer cases and deaths, United States, 2024.

| Model | New Cases | Deaths |
|---|---|---|
| Lung | 234,580 | 125,070 |
| Female breast | 313,510 | 42,250 |
| **Colorectum** | 152,810 | 53,010 |
| Prostate | 299,010 | 35,250 |
| Stomach | 26,890 | 10,880 |
| Liver | 41,630 | 29,840 |
| Thyroid | 44,020 | 2,170 |
| Cervix uteri | 13,820 | 4,360 |

## 2. Related Work

This section reviews the relevant literature on polyp segmentation using both traditional machine learning and recent deep learning approaches.

### 2.1. Traditional Methods

Traditionally, computer-aided detection (CAD) systems have been the primary means of identifying anomalies in wireless capsule endoscopy (WCE) images. The work of medical professionals is facilitated by these devices, which assist in automatically identifying problems such as tumors, polyps, and ulcers [7]. The majority of earlier CAD techniques relied on fundamental picture properties, including texture analysis, form measurement, and image segmentation into smaller areas [8, 9]. However, these techniques can lead to inaccurate or missed polyp detections since polyps frequently resemble the surrounding tissue. Additionally, methods that integrate feature analysis and adaptive clustering, as highlighted in recent literature, have demonstrated substantial potential for complex medical imaging tasks by incorporating morphological operations. [10]

### 2.2. Deep Learning-based Method

Deep learning has revolutionized polyp segmentation, surpassing traditional methods in accuracy and efficiency [11, 12]. Fully convolutional neural networks (CNNs) were first used for polyp segmentation by Akbari et al., who demonstrated notable advancements over traditional methods [13, 14]. Building on this foundation, recent work has focused on using more complex architectures, like ResUNet++ and UNet++, to handle the particular problems that polyp characteristics present [15, 16, 17]. In leveraging transfer learning, advanced classification models have demonstrated enhanced capabilities in distinguishing complex medical imaging features, which can be particularly useful for tasks such as brain tumor detection and classification, aligning with findings in other domains such as polyp segmentation and colorectal cancer diagnosis. Several new methods have been proposed to capture fine details about the boundary while avoiding the camouflage effect commonly seen in polyps [18]. XGBoost has also been utilized in medical imaging tasks for its robust classification capabilities, notably in segmenting and classifying regions of interest, including its successful application in ovarian cyst diagnosis, where it enhanced detection accuracy by combining feature extraction with efficient decision trees [19]. Several new methods have been proposed to capture fine details about the boundary while avoiding the camouflage effect commonly seen in polyps. Reverse attention modules have been introduced by PraNet [20]; they take into account the boundary cues and combine them with the global features that are extracted using a parallel partial decoder. To differentiate between regions that are polyp and those that are not, a boundary constraint network with a bilateral boundary extraction module is used. It contributed to a transformer-based camouflage identification module that was able to extract low-level feature-based hidden polyp cues. Despite advances in Deep Learning-based approaches such as TranNetR, which uses ResNet50 as an encoder, there is a need for more accurate and robust polyp segmentation models to address the challenges. It will take more investigation and advancement in this area to address these issues and raise the efficiency of polyp segmentation models.

### 2.3. Transformer-based Methods

The success of polyp segmentation based on transformer methods will lead to the development of more advanced technologies (Table 2). PraNet was a pioneer methodology, taking advantage of transformers' inherent strengths, and exploiting them for highly accurate, efficient, and accurate segmentation. In order to enhance efficiency and accuracy in polyp segmentation, ColonFormer [21] combined a hierarchical transformer with a pyramid network and incorporated an axial attention module. Enhanced multi-level attention mechanisms have been effectively employed in lightweight models to boost the identification accuracy of plant diseases, demonstrating the potential of attention-based architectures in improving classification performance [22]. Besides polyp segmentation, the broader field of medical image segmentation has prompted developments in architectures and techniques, resulting in promising results in radiology image segmentation. A few noteworthy developments are hierarchical transformers like ColonFormer [23] for capturing multi-scale features, which are important for the accurate segmentation of polyps of different sizes and shapes, and inherent attention mechanisms of transformers that make them quite good at detection and segmentation even in difficult conditions. Some of these techniques even used the combination of CNNs with transformers for local feature extraction and global context modeling. The integration of top-performing Transfer Learning (TL) architectures, by combining their strengths, can further improve the segmentation of medical images by enhancing feature extraction and classification accuracy [24]. The integration of a soft attention mechanism within CNN architectures, alongside the aggregation of features from all convolutional layers, has demonstrated significant improvements in accuracy and robustness in medical imaging tasks [25]. These new advancements help not only in tackling the variability of polyp appearance but are also helpful in making robust computer-aided diagnosis systems, which are central to early colorectal cancer diagnosis and consequent improvement in screening and treatment outcomes apart from motivating further research in this direction.

**TABLE 2** Comparative Analysis of Polyp Segmentation Models.

| Model | Methodology | Key Contributions | Key Findings/Performance |
|---|---|---|---|
| U-Net [26] | Encoder-decoder architecture with skip connections | Established the foundation for medical image segmentation with a simple and effective architecture | Effective for general segmentation tasks but limited in handling complex polyp structures |
| U-Net++ [27] | Nested U-Net with dense skip pathways | Improved boundary precision through enhanced skip connections | Higher boundary accuracy than U-Net but struggles with small, intricate details |
| ResU-Net++ [28] | U-Net++ enhanced with residual connections | Increases feature propagation efficiency, reducing gradient vanishing issues | Improved mDSC for large polyps but limited generalizability on varied polyp shapes |
| HarDNet-MSEG [29] | CNN with HarDNet backbone for multi-scale feature extraction | Combines efficiency of HarDNet with multi-scale capability for segmentation | Enhanced speed and accuracy, but less effective on small-scale segmentation details |
| ColonSegNet [30] | Lightweight CNN for colon polyp segmentation | Designed for efficient polyp segmentation with minimal computational load | Achieves decent results in real-time but lacks robustness on complex boundaries |
| UACANet [31] | Attention-based network with multiple attention blocks | Utilizes attention mechanisms to capture fine-grained features for improved polyp segmentation | High recall and precision, effective at fine detail, but computationally intensive |
| UNeXt [32] | CNN-based architecture with simple yet effective design | Emphasizes efficiency and simplicity for practical deployment | Lightweight and fast, but lower accuracy in complex segmentation tasks |
| TransNetR [33] | Transformer-based network with residual learning | Leverages transformers for better long-range dependencies in medical images | Achieves high precision and recall, especially for challenging segmentation tasks |
| Polyp-PVT [34] | Pyramid Vision Transformer model for polyp segmentation | Employs PVT backbone for multi-scale, attention-based feature extraction | High precision and mDSC, especially effective on in-distribution data |

## 3. Model Architecture

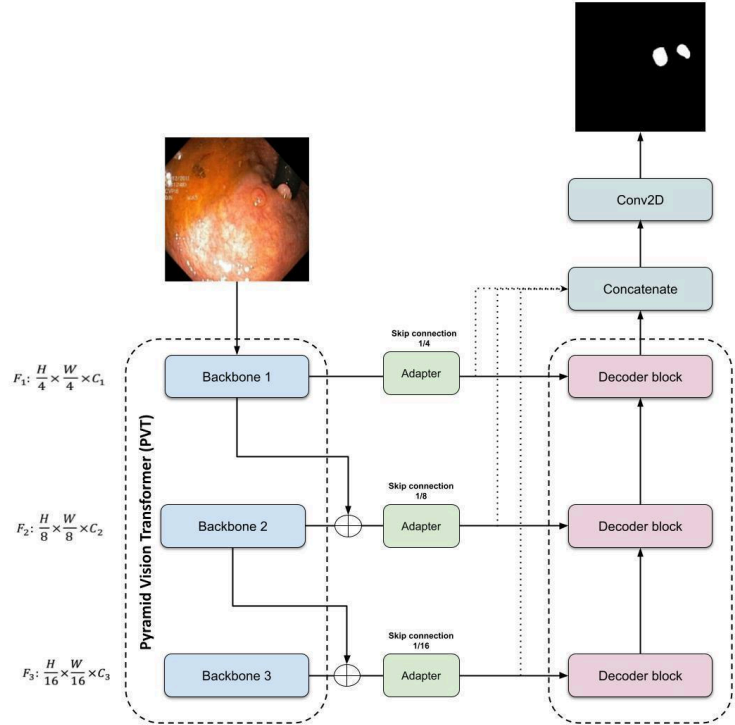

**FIGURE 1** PVTAdpNet (pyramid vision transformer Adapter network)

### 3.1. Model Overview

The architecture of PVTAdpNet is basically designed with a U-Net style encoder-decoder configuration, which is one of the most frequently used approaches for semantic segmentation (Figure 1).

In the proposed image segmentation model architecture, three Pyramid Vision Transformer backbones are used, all generating feature maps with different spatial dimensions and channels. These feature maps are down-sampled to have the same spatial dimensions and channels. The first backbone is down-sampled to the size of the second and then summed together, and similarly, the second backbone is downsampled to the size of the third and summed together. PVTAdpNet Decoder consists of a successive residual convolutional block with squeeze-and-excitation (SE) attention models. The residual connections are used to allow the smooth flow of gradients during the training process, while the SE blocks adapt channel-wise features. All of these will be further integrated using skip connections from the encoder through an adapter layer combined with concatenation at each stage to get localization cues. Finally, a $1 \times 1$ convolution predicts the segmentation mask output. In the following sections, residual blocks and SE attention are introduced, which are part of the PVTAdpNet decoder.

## 3.2. PVT Encoder

Polyp images often contain significant noise due to a variety of uncontrollable acquisition factors, such as motion blur, reflection, and rotation. Recent studies [35], [36] have demonstrated that vision transformers can achieve much better performance and be more robust to input distortions compared with traditional CNNs [37], [34], [38]. To leverage these advantages, we adopt a vision transformer as the backbone of our polyp segmentation model to extract robust and discriminative features.

We build upon the Pyramid Vision Transformer (PVT) [34] architecture, which efficiently reduces computational costs through spatial reduction attention. Our proposed model is flexible, allowing for the integration of different transformer backbones. In this work, we employ the enhanced PVTv2 [38] to further improve feature extraction. To adapt the model to polyp segmentation, we replace the final classification layer with a segmentation head that processes multi-scale feature maps from different network stages. These features capture both detailed appearance information and high-level semantic context, essential for accurate polyp segmentation.

Compared to CNN-based encoders, the PVT encoder has the ability to capture both detailed appearance information and high-level semantic context, which is essential for accurate polyp segmentation. This capability sets PVTAdpNet apart from models relying on CNN-based encoders (such as U-Net and UNet++), which may miss subtle boundary details or struggle to generalize across diverse polyp shapes and textures (Figure 2).

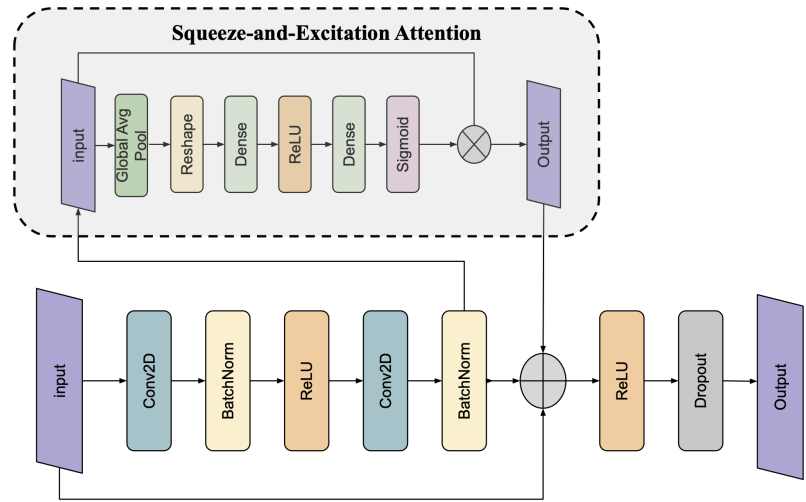

**FIGURE 2** PVTAdpNet Residual block structure.

### 3.3. Residual Learning

In PVTAdpNet, we employ residual learning, inspired by networks such as ResNet [39], to improve multi-scale feature aggregation within the decoder. Residual learning adds skip connections between network layers, enabling gradient flow and reducing training complexity in deep networks. For a typical convolutional block, the output at the $(l + 1)$-th block is derived from the previous block's output $x_l$ as follows:

$$x_{l+1} = H_l(x_l) \quad (1)$$

Here, $H_l$ represents the layer transformation. In contrast, a residual block incorporates an additive skip connection, allowing the output to bypass the block transformations:

$$x_{l+1} = H_l(x_l) + x_l \quad (2)$$

This simple modification provides several advantages:

- Gradient Flow Enhancement: The skip connection allows gradients to flow directly through shortcut paths, facilitating deeper network training.

- Feature Reuse: Later layers can reuse features from earlier layers, preserving crucial details for fine-grained tasks like segmentation.

- Focus on Novel Information: The model can focus on learning novel information rather than re-mapping the input at each layer.

In PVTAdpNet, residual learning supports effective multi-scale feature aggregation by combining information across different levels of the encoder-decoder hierarchy, thus improving segmentation details without significantly increasing model complexity. Our residual block is structured as follows:

1. Channel Reduction: A 1×1 convolution reduces the dimensionality of feature maps.
2. Feature Extraction: Two 3×3 convolutions are applied for detailed feature extraction.
3. Squeeze-and-Excitation (SE) Attention: Enhances channel-wise feature response (detailed below).
4. Shortcut Addition: Element-wise addition of the input through the shortcut path.
5. Non-linearity: ReLU activation introduces non-linearity after feature aggregation

This configuration allows PVTAdpNet to retain localization cues for polyp segmentation, leveraging residual connections to handle complex feature representations effectively.

### 3.4. Squeeze-and-Excitation Attention

The SE attention mechanism is a crucial part of PVTAdpNet's architecture, particularly within the residual blocks. SE attention operates by learning to recalibrate channel-wise feature responses, allowing the model to emphasize relevant features and reduce redundancy. The SE block operates in two stages: Squeeze and Excitation [37].

1. Squeeze: The feature map $U \in R^{H \times W \times C}$ undergoes global average pooling to produce a channel descriptor $z \in R^c$, summarizing each channel's global information [40]. This is computed as:

$$z_c = \frac{1}{H \times W} \sum_{i=1}^{H} \sum_{j=1}^{W} u_c(i, j) \quad (3)$$

2. Excitation: To capture channel dependencies, $z$ passes through a two-layer fully connected bottleneck with a ReLU activation and a final sigmoid, generating adaptive weights $s \in R^c$:

$$s = \sigma(W_2 \delta(W_1 z)) \quad (4)$$

where $W_1$ and $W_2$ are learnable parameters.

3. Channel Reweighting: The recalibrated feature map $\tilde{U}$ is obtained by scaling each channel of $U$ with the corresponding weight in $s$:

$$\tilde{U} = s_c \times u_c \quad (5)$$

In contrast to standard attention mechanisms, SE Attention allows the model to retain critical localization cues necessary for accurate polyp segmentation. When combined with residual connections, SE Attention enables the model to effectively manage complex feature representations, achieving higher accuracy in segmenting small or less prominent polyps than other models that lack this targeted attention refinement [41].

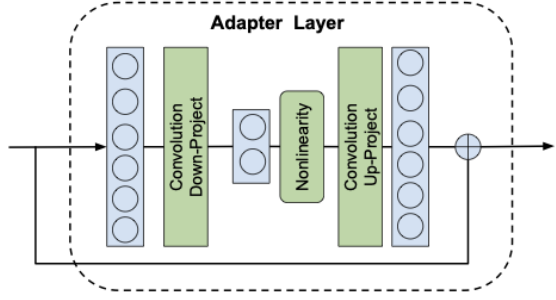

**FIGURE 3** PVTAdpNet Adapter block structure

### 3.5. Adapter base PVT

Adapters, as proposed in [42], have gained significant traction in the field of natural language processing (NLP) for the fine-tuning of large language models. By introducing specialized modules within transformer encoders, adapters allow pre-trained models to be quickly adapted to a wide range of downstream NLP tasks. Beyond NLP, the concept of adapters has found applications in computer vision, where it has been employed for incremental learning [43] and domain adaptation [44], demonstrating its versatility across different domains. for having powerful dense prediction using adapters for PVT (Figure 3). The parallel adapter can be formulated below where $H_i$ is the input of a specific layer:

1. Down-Projection: Reduce the dimensionality of $H_i$ using a convolutional layer $W_{down}$ where $H_d$ is the down-projected feature map:

$$H_d = H_i \cdot W_{down} \quad (6)$$

   This step lowers computation by reducing feature map size.

2. Non-linearity: Apply an activation function (e.g., ReLU, Leaky ReLU, or GELU) to introduce non-linearity, enhancing the model's ability to capture complex patterns where $H_a$ is the activated feature map:

$$H_a = f(H_d) \quad (7)$$

3. Up-Projection: Restore the feature map to its original dimensions using another convolutional layer $W_{up}$ where $H_d$ is the up-projected feature map:

$$H_{up} = H_a \cdot W_{up} \quad (8)$$

4. Parallel Adapter Integration: Inspired by the concept of Parallel Adapters [45], add a parallel transformation pathway alongside the main adapter. This structure refines output by introducing additional learnable pathways where $H_o$ is the output of a specific layer:

$$H_o = H_{up} + f(H_i W_{down}) W_{up} \quad (9)$$

Unlike traditional or single-path transformation layers, the Adapter layer in PVTAdpNet enables efficient transformation and feature recalibration, which is crucial for dense prediction tasks like polyp segmentation. By utilizing both main and parallel pathways, the model achieves enhanced flexibility, adapting effectively to diverse input features and producing robust segmentation outcomes. This configuration distinguishes PVTAdpNet from simpler models, where single-path transformations may limit the model's ability to handle diverse and complex patterns in clinical imagery.

## 4. Experiment

### 4.1. Loss Function

To optimize PVTAdpNet training, we employed a combined loss function incorporating both Jaccard and Dice components. The Jaccard loss quantifies the overlap between predicted and ground truth segmentations, providing a measure of segmentation quality. It is defined as:

$$L_{Jaccard}(y, \hat{y}) = \alpha \times (1 - \frac{\sum_{c}^{C}(y_c \times \hat{y}_c) + \alpha}{\sum_{c}^{C}(y_c + \hat{y}_c - y_c \times \hat{y}_c) + \alpha}) \qquad (10)$$

The segmentation process is enhanced by employing a loss function that effectively regulates model performance based on tissue characteristics [46]. The Jaccard loss, synonymous with the Intersection over Union (IoU) metric, quantifies the overlap between predicted $\hat{y}$ and ground truth segmentations These two labels are demonstrated in the one-hot vector to present classes C being their length. However, to prevent the exploding gradient, there is a smoothing factor called alpha α, which helps stabilize the training result [47].

Dice loss is derived from the Dice coefficient, a metric that assesses the similarity between predicted and ground truth segmentations. It is calculated as the ratio of twice the intersection of the two sets to the sum of their individual sizes. Dice loss is defined as:

$$L_{Dice}(X, Y) = 1 - \frac{2 \times |X \cap Y| + \epsilon}{|X| + |Y| + \epsilon} \qquad (11)$$

X is the predicted segmentation, Y is the ground truth segmentation (set of ground truth pixels), |X∩Y| represents the size of the intersection between the predicted and ground truth sets., |X| and |Y| are the sizes (or sums) of the predicted and ground truth sets, respectively. ε is a small constant added to avoid division by zero and to stabilize the loss function.

The total loss is a weighted sum of binary cross-entropy loss, Dice loss, and Jaccard loss:

$$L_{Total} = L_{BCE} + L_{Dice} + L_{Jaccard} \qquad (12)$$

## 4.2. Evaluation Metrics

We used most of the evaluation metrics applied in recent studies on image segmentation. Specifically, we used the Weighted F-measure ($F_{\beta}^{\omega}$), mDice, and mIoU. This set of metrics gives a good all-round assessment of how well the model is performing with regard to segmentation accuracy.

**4.2.1. Weighted F-measure**: Combines precision and recall into a single metric, while β controlling their relative importance. It is defined as:

$$F_{\beta}^{\omega} = \frac{(1 + \beta^2) \times Precision^{\omega} \times Recall^{\omega}}{\beta^2 \times Precision^{\omega} + Recall^{\omega}} \qquad (13)$$

where $Precision^{\omega}$ represents weighted precision and $Recall^{\omega}$ denotes weighted recall.

**4.2.2. Mean Dice Coefficient**: Measures The Overlap between the predicted segmentation and ground truth masks. It is defined as:

$$D_{Dice}(X, Y) = \frac{2 \times |X \cap Y|}{|X| + |Y|} = \frac{2 \times TP}{2 \times TP + FP + FN} \qquad (14)$$

where $X$ represents the predicted segmentation, $Y$ is the ground truth, $TP$ denotes true positives or the common area of two masks X and Y, $FP$ indicates false positives, and $FN$ marks false negatives.

## 4.3. Datasets

To establish a robust benchmark for polyp segmentation, we leveraged three publicly accessible datasets: Kvasir-SEG [48], CVC-ClinicDB [49], and PolypGen [50]. These datasets collectively provide a diverse range of polyp morphologies, image qualities, and clinical scenarios, enabling a comprehensive evaluation of our proposed method (Table 3).

Table 3 Comparison of Polyp Segmentation Datasets

| Feature | Kvasir-SEG | CVC-ClinicDB | PolypGen |
|---|---|---|---|
| Number of Images | 1000 | 612 | 1537 |
| Data Sources | Routine clinical procedures | routine clinical procedures | six different medical centers |
| polyp Diversity | size, shape, texture, color | size, shape, texture, color (especially small and large) | size, shape, texture, color, diverse, patient populations |
| Segmentation Masks | Accurate, precise, polyp boundaries | Accurate, precise, polyp boundaries | Accurate, precise, polyp boundaries |
| other Features | High diversity, accurate masks, clear split | clinical relevance, diverse morphology | large scale, high diversity, high quality |

**Kvasir-SEG:** This dataset contains 1,000 endoscopic images that are all equipped with pixel-wise segmentation masks annotated by medical experts. There is a very wide variety regarding the size, shape, and texture of polyps, corresponding to their real-world clinical spectrum.

**CVC-ClinicDB** Presents a dataset of 612 endoscopic images together with their corresponding segmentation masks. The database provides a clinically representative dataset from 31 colonoscopy sequences, featuring a comprehensive capture of polyp appearance variability—both small and large polyps.

**PolypGen:** PolypGen is a large-scale dataset of 1,537 endoscopic images with their segmentation masks. Its data is drawn from six medical centers, thereby casting a wide net over patient populations and polyp characteristics (Figure 4). Being a large and diverse dataset, PolypGen significantly contributes to the development of robust models for this problem setting.

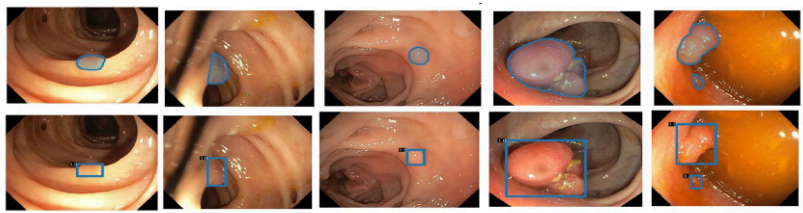

(a) C1: Ambroise Pare Hospital, Paris, France

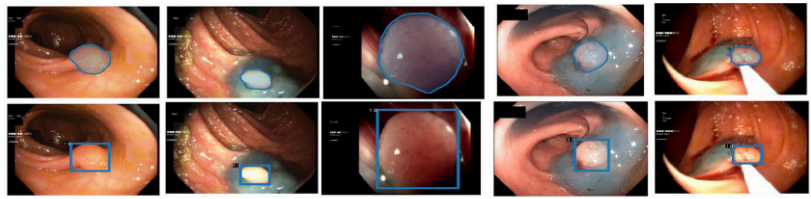

(b) C2: Istituto Oncologico Veneto, Padova, Italy

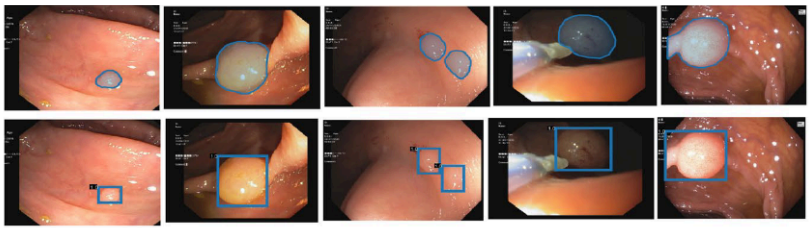

(c) C3: Centro Riferimento Oncologico, IRCCS, Italy

**FIGURE 4** Samples PolypGen dataset

*4.4. Experiment setup and configuration*

All models in our study utilize the Kvasir-SEG dataset, consisting of 1000 image-mask pairs. We train our models on 880 pairs, reserving the rest for validation and testing. To enhance the dataset, we apply extensive data augmentation techniques. The experiments were executed using an NVIDIA RTX 3090 GPU within the PyTorch environment. The models are optimized with the Adam optimizer, set at a learning rate of 1e-4 and a batch size of 8. For loss calculation, a combination of binary cross-entropy and dice loss is employed to improve training effectiveness (Table 4).

**TABLE 4** Hyperparameters of proposed method

| Hyperparameter | Value |
|---|---|
| Learning Rate | 1e-4 |
| Optimizer | Adam |
| Early Stopping | Yes, Patience = 5 |

|  | The head architecture | Conv2D, 4:1:256:256 |
|---|---|---|
|  | Number of epochs | 30 |
|  | Loss | DiceJaccardLoss |
|  | Batch Size | 4 |

## 5. Result

In this subsection, we evaluate the performance of PVTAdpNet in comparison to other state-of-the-art methods of polyp segmentation, including U-Net, U- Net++, ResU-Net++, HarDNet-MSEG, ColonSegNet, UACANet, UNeXt, TransNetR. These methods span a wide scope of architectural designs from CNN-based to transformer-based models and have proved to be competitive for polyp segmentation tasks.

**TABLE 5** Quantitative results on the Kvasir-SEG test dataset.
Training dataset: Kvasir-SEG – Test dataset: Kvasir-SEG

| Method | mIoU | mDSC | Rec. | Prec. | F2 |
|---|---|---|---|---|---|
| U-Net [23] | 0.7472 | 0.8264 | 0.8504 | 0.8703 | 0.8353 |
| U-Net++ [24] | 0.7420 | 0.8228 | 0.8437 | 0.8607 | 0.8295 |
| ResU-Net++ [24] | 0.5341 | 0.6453 | 0.6964 | 0.7080 | 0.6576 |
| HarDNet-MSEG [26] | 0.7459 | 0.8260 | 0.8485 | 0.8652 | 0.8358 |
| ColonSegNet [27] | 0.6980 | 0.7920 | 0.8193 | 0.8432 | 0.7999 |
| UACANet [28] | 0.7692 | 0.8502 | 0.8799 | 0.8706 | 0.8626 |
| UNeXt [29] | 0.6284 | 0.7318 | 0.7840 | 0.7656 | 0.7507 |
| TransNetR [30] | 0.8016 | 0.8706 | 0.8843 | 0.9073 | 0.8744 |
| Polyp-PVT [31] | 0.8640 | 0.917 | 0.9260 | 0.9180 | 0.9110 |
| **PVTAdpNet (Ours)** | **0.8726** | **0.9212** | **0.9380** | **0.9280** | **0.9283** |

It can be seen that the quantitative results on the Kvasir-SEG dataset show that PVTAdpNet achieves superior scores in most of the metrics (Table 5). The results further prove the effectiveness of PVTAdpNet in polyp segmentation, especially for some key metrics. With a Dice of 0.9212, mIoU of 0.8726, and F2 of 0.9283 The strength of PVTAdpNet lies in its ability to maintain a good balance between performance and efficiency and PVTAdpNet's architecture design in achieving superior results across most evaluation measures on the Kvasir-SEG dataset.

**TABLE 6** Training dataset: Kvasir-SEG – Test dataset: CVC-ClinicDB

| Method | mIoU | mDSC | Rec. | Prec. | F2 |
|---|---|---|---|---|---|
| U-Net [23] | 0.5433 | 0.6336 | 0.6982 | 0.7891 | 0.6563 |
| U-Net++ [24] | 0.5475 | 0.6350 | 0.6933 | 0.7967 | 0.6556 |
| ResU-Net++ [25] | 0.3585 | 0.4642 | 0.5880 | 0.5770 | 0.5084 |
| HarDNet-MSEG [26] | 0.6058 | 0.6960 | 0.7173 | 0.8528 | 0.7010 |
| ColonSegNet [27] | 0.5090 | 0.6126 | 0.6564 | 0.7521 | 0.6246 |
| UACANet [28] | 0.6808 | 0.7659 | 0.7639 | 0.8820 | 0.7599 |
| UNeXt [29] | 0.3901 | 0.4915 | 0.6125 | 0.6609 | 0.5318 |
| TransNetR [30] | 0.6912 | 0.7655 | 0.7571 | **0.9200** | 0.7565 |
| Polyp-PVT [31] | **0.727** | **0.808** | 0.7571 | 0.9190 | 0.795 |
| **PVTAdpNet (Ours)** | 0.6989 | 0.7891 | **0.8691** | 0.7874 | **0.8217** |

**TABLE 7** Training dataset: Kvasir-SEG – Test dataset: PolypGen (C1)

| Method | mIoU | mDSC | Rec. | Prec. | F2 |
|---|---|---|---|---|---|
| U-Net [23] | 0.5772 | 0.6469 | 0.6780 | 0.8464 | 0.6484 |
| U-Net++ [24] | 0.5857 | 0.6611 | 0.6953 | 0.8247 | 0.6700 |
| ResU-Net++ [25] | 0.4204 | 0.5239 | 0.6390 | 0.5789 | 0.5557 |
| HarDNet-MSEG [26] | 0.6256 | 0.7121 | 0.7800 | 0.7933 | 0.7344 |
| ColonSegNet [27] | 0.5514 | 0.6386 | 0.7130 | 0.7423 | 0.6551 |
| UACANet [28] | 0.6386 | 0.7189 | 0.7553 | 0.8476 | 0.7254 |
| UNeXt [29] | 0.4481 | 0.5386 | 0.6421 | 0.6912 | 0.5686 |
| TransNetR [30] | 0.6538 | 0.7204 | 0.7438 | **0.8778** | 0.7269 |
| **PVTAdpNet (ours)** | **0.7591** | **0.8320** | **0.9072** | 0.8181 | **0.8593** |

**TABLE 8** Training dataset: Kvasir-SEG – Test dataset: PolypGen (C2)

| Method | mIoU | mDSC | Rec. | Prec. | F2 |
|---|---|---|---|---|---|
| U-Net [23] | 0.5702 | 0.6338 | 0.7347 | 0.7368 | 0.6495 |
| U-Net++ [24] | 0.5612 | 0.6240 | 0.7189 | 0.7631 | 0.6383 |
| ResU-Net++ [25] | 0.2779 | 0.3431 | 0.5003 | 0.4198 | 0.3606 |
| HarDNet-MSEG [26] | 0.5667 | 0.6311 | 0.7267 | 0.7149 | 0.6376 |
| ColonSegNet [27] | 0.4659 | 0.5371 | 0.6443 | 0.6789 | 0.5439 |
| UACANet [28] | 0.6091 | 0.6887 | 0.8540 | 0.6870 | 0.7222 |
| UNeXt [29] | 0.3780 | 0.4583 | 0.6373 | 0.5239 | 0.4837 |
| TransNetR [30] | 0.6608 | 0.7232 | 0.8071 | 0.8096 | 0.7366 |
| **PVTAdpNet (ours)** | **0.7384** | **0.7953** | **0.8951** | **0.8122** | **0.8097** |

**TABLE 9** Training dataset: Kvasir-SEG – Test dataset: PolypGen (C3)

| Method | mIoU | mDSC | Rec. | Prec. | F2 |
|---|---|---|---|---|---|
| U-Net [23] | 0.6769 | 0.7481 | 0.7637 | 0.8787 | 0.7518 |
| U-Net++ [24] | 0.6530 | 0.7254 | 0.7526 | 0.8568 | 0.7332 |
| ResU-Net++ [25] | 0.3786 | 0.5109 | 0.6463 | 0.5484 | 0.5545 |
| HarDNet-MSEG [26] | 0.6623 | 0.7440 | 0.7947 | 0.8180 | 0.7619 |
| ColonSegNet [27] | 0.6181 | 0.7064 | 0.7520 | 0.7907 | 0.7221 |
| UACANet [28] | 0.7074 | 0.7870 | 0.7954 | 0.8893 | 0.7877 |
| UNeXt [29] | 0.4654 | 0.5534 | 0.6265 | 0.6868 | 0.5740 |
| TransNetR [30] | 0.7217 | 0.7874 | 0.7904 | **0.9133** | 0.7863 |
| **PVTAdpNet (ours)** | **0.8167** | **0.8851** | **0.9197** | 0.8778 | **0.9010** |

The excellent performance of PVTAdpNet on the CVC-ClinicDB and polypGen datasets emphasizes its robustness and generalization ability (Tables 6, 7, 8 and 9). The obtained results are a mDice of 0.7891, and mIoU of 0.6989 on CVC-ClinicDB, mDice of 0.8320, mIoU 0.7591 C1 PolypGen dataset, mDice of 0.7953 mIoU 0.7384 on C2 PolypGen dataset and mDice of 0.8851, mIoU 0.8167 C3 PolypGen dataset That is to say, all the evaluation metrics in this study could achieve high scores most of the time by the model, which proved its potential for reliable and accurate segmentation of the polyp in the clinical setting. Its success across multiple measures proves its adaptability and reliability, hence making it a strong candidate for its wide adoption in clinical practice.

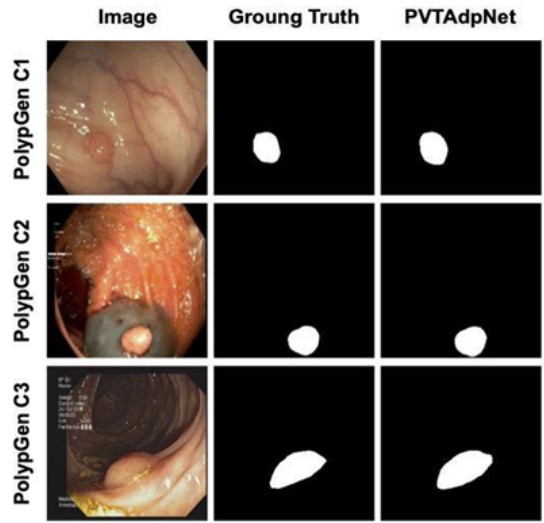

(a) Qualitative result on PolypGen dataset

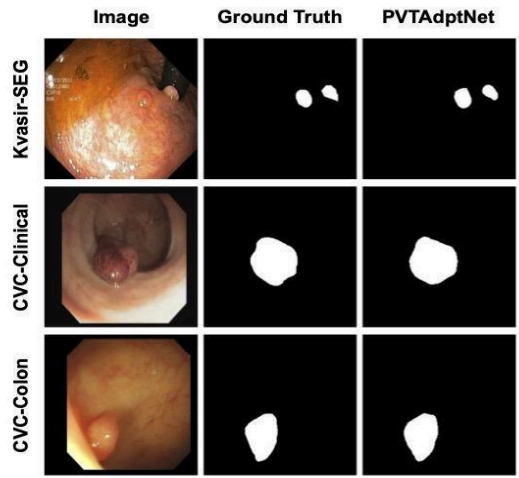

(b) Qualitative result on Kvasir-SEG, CVC-ClinicDB datasets

**FIGURE 5** Qualitative results of PVTAdpNet on sample images.

Qualitative result for sample images from datasets Kvasir-SEG, CVC-ClinicDB, and PolypGen. The resulting segmentation maps, generated by the proposed PVTAdpNet, are highly close to the ground-truth annotated maps, with the polyp areas' boundaries clearly defined. The model shows good generalization capabilities; hence, it does a good segmentation of polyps, even when they have different sizes, shapes, and appearances. More specifically, in the PolypGen dataset, it can be seen that there is much clarity that PVTAdpNet can lead to highly fine segmentation for different types of polyps, revealing bright and neat segmentation maps (Figure 5a). At the same time, the performance of good accuracy was retained in delineating polyp boundaries, even for the most difficult cases and a great variety of polyp presentations, in both Kvasir-SEG and CVC-ClinicDB datasets (Figure 5b). Such performance on different datasets may, in a sense, provide some clinical reliability regarding polyp segmentation applicability.

# 6. Ablation Studies

PVTBaseNet has a U-Net-style encoder-decoder architecture (Figure 6), commonly used for semantic segmentation [26]. The basic Pyramid Vision Transformer (PVT) model serves as the encoder, while the decoder comprises two simple convolution layers in each block [34]. Skip connections are also incorporated, including a 1×1 Convolution layer, Batch Normalization, and a ReLU activation function (CBR), which are upsampled and concatenated with the corresponding decoder block outputs to enhance feature propagation and improve segmentation accuracy [51, 52]. Additionally, residual blocks are integrated into the architecture to further refine feature maps, followed by an upsampling layer and a concatenation step to combine multi-scale features. This design enables the model to capture and utilize low-level details alongside high-level semantic information, leading to more precise segmentation results. The final output layer applies a sigmoid activation function to produce the segmentation mask.

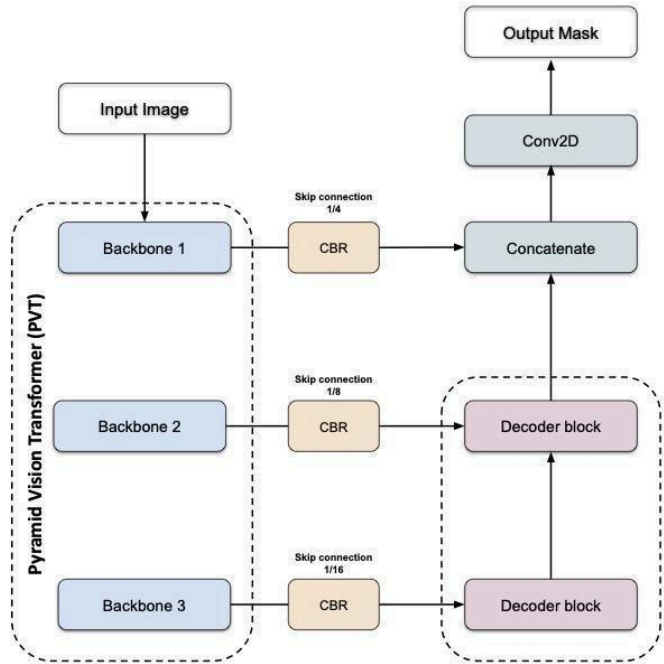

**FIGURE 6** PVTBaseNet (pyramid vision transformer basic network)

To investigate the impact of various components within the PVTAdapNet model, we conducted comprehensive ablation studies using the Kvasir-SEG dataset We evaluated the performance of three variants of PVTAapNet:

**TABLE 10** Ablation studies test on the Kvasir-SEG dataset.

| Methods | mIoU | Rec. | Prec. | F2 |
| --- | --- | --- | --- | --- |
| TransNetR | 0.8016 | 0.8843 | 0.9073 | 0.8744 |
| **PVTBaseNet** | 0.8445 | 0.9195 | 0.9170 | 0.9048 |
| **PVT + DS Enc** | 0.8551 | 0.9263 | 0.9227 | 0.9110 |
| **PVT + DS Enc + Res** | 0.8526 | 0.9015 | **0.9444** | 0.9096 |
| **PVTAdpNet** | **0.8726** | **0.9380** | 0.9280 | **0.9283** |

The results of our erosion studies on the PVTAdpNet model can be seen in terms of the contribution of each component (Table 10). In particular, we removed some components one by one, measuring segmentation performance in terms of various metrics, such as mean Intersection over Union, recall, precision, and the F2 score.

1. **PVT + DS Enc:** This variant builds upon PVTBaseNet by integrating a downsampling encoder block (DS Enc). This enhancement leads to improved metrics, with the model achieving an mIoU of 0.8551, recall of 0.9263, precision of 0.9227, and an F2 score of 0.9110.
2. **PVT + DS Enc + Res:** The PVT + DS Enc + Res model further augments the PVT + DS Enc architecture by adding improved residual connections (Res). This model demonstrates slightly varied performance metrics, achieving an mIoU of 0.8526, recall of 0.9015, precision of 0.9444, and an F2 score of 0.9096.
3. **PVTAdpNet:** The PVTAdpNet model represents our most advanced architecture, which incorporates an adapter layer into the skip connection of the PVT + DS Enc + Res model. This model demonstrates superior performance across all evaluated metrics, achieving an mIoU of 0.8726, recall of 0.9380, precision of 0.9280, and an F2 score of 0.9283.

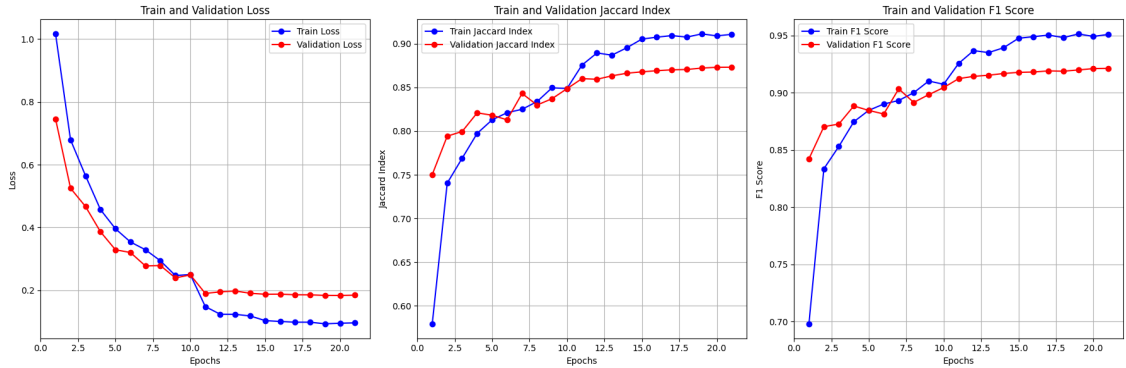

(a) Using the adapter in skip-connection (PVTAdapNet) using evaluation metrics

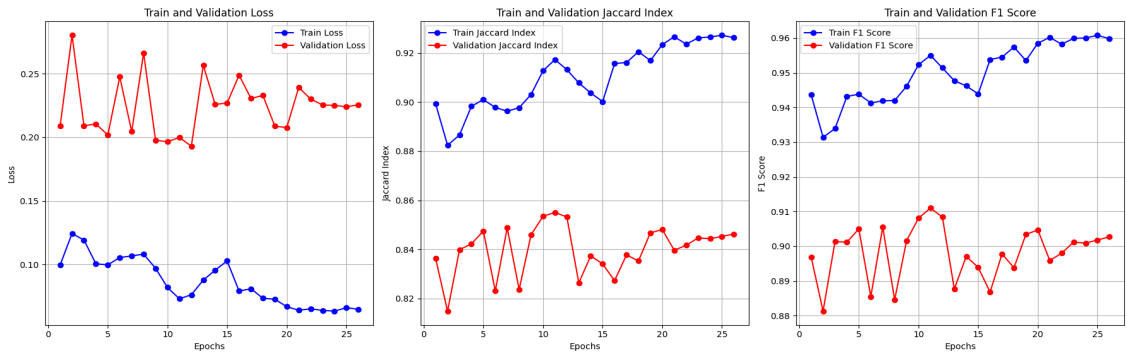

(b) Adding DownSampling to the Encoder (DS Enc) model architecture using evaluation metrics

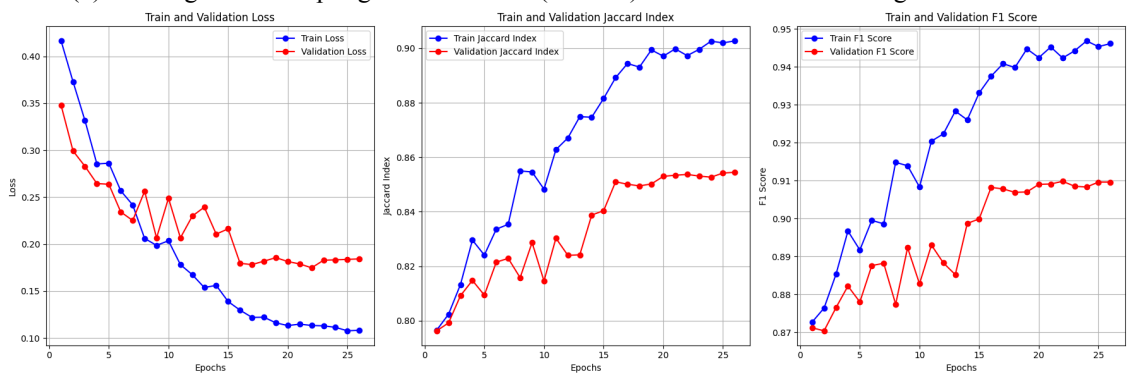

(c) Adding HyperRes to the model architecture using evaluation metrics

**FIGURE 7** Comprehensive Evaluation of Model Enhancements Using DownSampling, HyperRes, and Skip-Connection Adapter (PVTAdapNet) Based on Loss, Jaccard Index, and F1 Score Across Epochs

Among the three models, the PVTAdpNet performs most robustly, an effect obviously visible in its rapid convergence of training loss and its value, which is very close to the validation loss, thus proving efficient learning with minimal overfitting. In the case of PVTAdpNet, there are trends toward stable validation with a high Jaccard Index and F1 score, which are very near to the corresponding training metrics, reflecting superior generalization and accuracy (Figure 7a). On the other hand, PVT + DS Enc works well with stable validation metrics and efficient learning, but slightly more fluctuation and less rapid convergence compared to PVTAdpNet (Figure 7b). Although improved, validation metrics for the Improved PVT model still have large fluctuations, hence possibly suffering from overfitting or high variability. Generally, the most efficient and generalizable model turns out to be PVTAdpNet, followed by PVT + DS Enc and the Improved PVT model (Figure 7c).

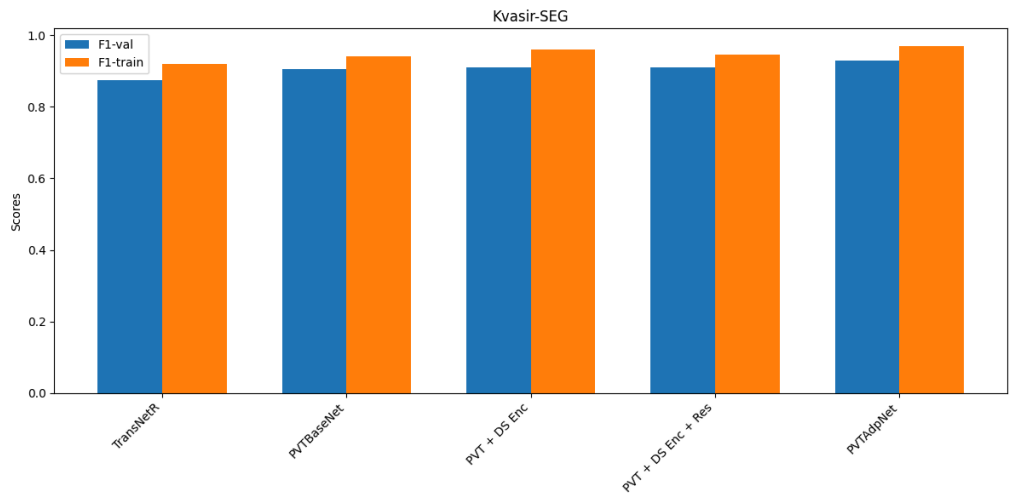

**FIGURE 8** Display Performance comparison of PVTAdpNet and other Ablation study methods on the Kvasir-SEG dataset using the F1-train and F1-val metric.

According to the results of the ablation study (Figure 8), From the bar plot, it is clear to see which component contributes to what extent to the segmentation performance. Although the full PVTAdpNet model achieves the highest score across all metrics, variants with only DownSampling in the Encoder, Residual connection, Adapter layer, and pre-training show notable performance improvement. This clearly visualizes the necessity of the components incorporated in the architecture of PVTAdpNet for accurate polyp segmentation through this ablation study.

## 7. Conclusion

Timely and accurate segmentation of polyps is very crucial for the early detection and prevention of colorectal cancer. Specifically, the authors propose a new deep-learning architecture called PVTAdpNet for overcoming some of the challenges in polyp segmentation from colonoscopy images. Combining transformer-based attention mechanisms into the residual upsampling framework with an adaptive encoder, PVTAdpNet can accurately model long-range dependencies, adapt to diversity in polyp appearance, and enhance feature extraction. Very challenging benchmark datasets prove that PVTAdpNet significantly outperforms current state-of-the-art methods in segmentation accuracy, especially for small and subtle polyps. It is lightweight, enabling real-time inference, hence quite suitable for clinical deployment. Further, ablation studies confirm that each component of PVTAdpNet makes contributions to the final performance due, especially very important transformer-based attention and adaptive encoders. The fact that PVTAdpNet has been successfully applied to polyp segmentation problems opens the window for new possibilities of computer-aided diagnosis in the screening of colorectal cancer and lays a foundation for further research into medical image segmentation. In future work, we would be interested in integrating a new encoding system as an alternative encoder for more feature extraction and efficiency.Furthermore, we hope to increase dataset diversity, new clinical applications, and extend PVTAdpNet to a wide range of medical imaging tasks for better early diagnosis and improved patient outcomes.

### Data availability

The code for the PVTAdpNet model is open-source and available on GitHub: https://github.com/ayousefinejad/PVTAdpNet.git. The datasets used in this study are publicly available and can be accessed from their respective sources: Kvasir-SEG (https://datasets.simula.no/kvasir-seg/), CVC-ClinicDB (https://polyp.grand-challenge.org/CVCClinicDB/), EndoScene (https://figshare.com/articles/figure/Polyp_DataSet_zip/_21221579), EndoTect (https://endotect.com/), PolypGen (https://github.com/DebeshJha/PolypGen)and Kvasir- Sessile (https://datasets.simula.no/kvasir-seg/).

### Acknowledgments

This work did not receive financial support from any funding agency.
The authors have no conflict of interest.